\documentclass[11pt,a4paper]{article}
\usepackage[hyperref]{acl2021}
\usepackage{times}
\usepackage{latexsym}
\usepackage{listings}
\usepackage{multirow}
\usepackage{amssymb}
\usepackage{float}
\usepackage{makecell}
\usepackage{microtype}
\usepackage{booktabs}
\usepackage{amsmath}
\usepackage{graphicx}
\aclfinalcopy 

\title{ASC analyzer: A Python package for \\ measuring argument structure construction usage in English texts}

\author{
  Hakyung Sung\textsuperscript{1,2} \and Kristopher Kyle\textsuperscript{1} \\
  \textsuperscript{1}Department of Linguistics, University of Oregon, Eugene, OR, USA \\
  \textsuperscript{2}Department of Psychology, Rochester Institute of Technology, Rochester, NY, USA \\
  \texttt{hksgla@rit.edu, kkyle2@uoregon.edu}
}

\begin{document}
\maketitle
\begin{abstract}
Argument structure constructions (ASCs) offer a theoretically grounded lens for analyzing second language (L2) proficiency, yet scalable and systematic tools for measuring their usage remain limited. This paper introduces the ASC analyzer, a publicly available Python package designed to address this gap. The analyzer automatically tags ASCs and computes 50 indices that capture diversity, proportion, frequency, and ASC-verb lemma association strength. To demonstrate its utility, we conduct both bivariate and multivariate analyses that examine the relationship between ASC-based indices and L2 writing scores.
\end{abstract}

\section{Introduction}

Linguistic complexity has long been recognized as an important construct in second language (L2) production research. It is commonly conceptualized along two complementary dimensions: absolute complexity and relative complexity \cite{bulte2012defining, bulte2025complexity}. Absolute complexity refers to the structural properties of language, where complexity increases with the number and interrelation of constituent units. In contrast, relative complexity pertains to the cognitive effort involved in using particular forms, typically operationalized via their relative frequency and the strength of their statistical contingencies. To date, a wide range of lexicogrammatical units have been proposed to quantify complexity dimensions, including argument structure constructions (ASCs).

ASCs are clausal-level lexicogrammatical patterns, each anchored by a main verb and a specific argument configuration (e.g., \citealp{goldberg1995constructions, goldberg2013argument, diessel2004acquisition, ellis2009constructing}). In L2 research, two main approaches have examined their linguistic complexity. One builds on Goldberg’s \citeyearpar{goldberg1995constructions} inheritance hierarchy, which organizes ASCs by semantic role complexity and posits that learners acquire them in a developmental sequence—from simpler constructions (e.g., simple transitives) to more complex ones (e.g., transitive resultatives). Empirical studies have operationalized this trajectory by analyzing the diversity or proportion of ASCs in learner texts (e.g., \citealp{hwang2023automatic, kim2023constructional}).

The other line of research focuses on the relationship between verbs and constructions. It posits that language learners initially tend to produce ASCs with semantically prototypical verbs (i.e., those that strongly instantiate verb-specific argument patterns), which gradually generalize into more abstract constructions \cite{ninio1999pathbreaking}. For instance, learners may first acquire the ditransitive construction using prototypical verbs like ``give'' (e.g., ``She \textbf{gave} him a book''), before extending it to less prototypical verbs such as ``offer'' or ``send''. This developmental trajectory has often been assessed using measures such as the relative frequency and statistical contingency between verbs and constructions (e.g., \citealp{ellis2009construction, kyle2017assessing}). While a growing body of empirical research has supported both developmental patterns (§ \ref{section2:1}), scalable and systematic tools for extracting and analyzing ASC-based indices remain underdeveloped.

To address this gap, we present ASC analyzer, a Python package that leverages a RoBERTa-based ASC tagger \cite{sung2024leveraging} trained on a gold-standard ASC treebank \cite{sung2024annotation}. The tool automatically labels ASCs and computes a suite of indices capturing their diversity, proportion, frequency, and ASC-verb lemma association strength. We also demonstrate the application of the tool through a sample analysis of 6,482 English learner essays from the ELLIPSE corpus \cite{crossley2023english}, examining the relationship between ASC-based indices and L2 English writing proficiency.

\section{Background}

\subsection{Empirical findings on ASC usage in L2 production}
\label{section2:1}
From a usage-based constructionist perspective, language is a network of form-meaning pairings (i.e., constructions) that emerge through repeated exposure and use \cite{fillmore1988mechanisms, goldberg1995constructions, langacker1987nouns}. The constructions develop from actual language use and are shaped by patterns of frequency, distribution, and co-occurrence in the input and output \cite{bybee2010language, diessel2015usage, ellis2012formulaic, stefanowitsch2003collostructions}. As learning is usage-driven, linguistic knowledge accumulates incrementally, shaped by each learner’s unique language experience. Empirical studies in this framework examine constructions at varying levels of granularity (e.g., words, phrases, clauses, discourse), with particular attention to clausal-level ASCs, which are schematic form–meaning pairings that encode core semantic relations such as motion, causation, and transfer \cite{goldberg1995constructions}.

A body of L2 research has illustrated how ASC usage can be investigated across different L2 modalities and proficiency scores. In L2 writing, for example, Hwang and Kim \citeyearpar{hwang2023automatic} found that more proficient L2 writers tend to produce a higher proportion of complex ASCs such as resultatives. Kim et al. \citeyearpar{kim2023constructional} further demonstrated that ASC-based indices outperform traditional T-unit measures in predicting writing proficiency. Another line of research highlights the role of verb-construction pairings. Kyle and Crossley \citeyearpar{kyle2017assessing} found that L2 essay scores were negatively correlated with the relative frequency of these pairings but positively correlated with their strength of association, suggesting that advanced learners favor less frequent but more strongly associated verb–construction combinations.

Although less studied, L2 speaking shows similar patterns. Choi and Sung \citeyearpar{choi2020utterance} found that ASC use (especially transitive constructions) explained most of the variance in L2 fluency. Kim and Ro \citeyearpar{kim2023assessment} reported that advanced L2 speakers produced a wider range of verb–construction combinations. A recent study by Sung and Kyle \citeyearpar{SungKyle2025} further confirmed these findings, showing that ASC-based indices alone accounted for 44\% of the variance in L2 oral proficiency scores.

\subsection{ASC tagger}
In this context, reliable identification of ASCs is essential for investigating their relationship with L2 proficiency in large-scale learner corpora \citep{kyle2023argument}. To meet this need, prior studies have explored a range of tagging methodologies, including dependency parsing \citep[e.g.,][]{o2010towards, romer2014second, kyle2017assessing}, rule-based approaches built on top of dependency structures \citep[e.g.,][]{hwang2023automatic, kim2023constructional}, and methods that leverage semantic role labels \citep[e.g.,][]{jeon2024corpus, kyle2023argument}. Of particular relevance, \citet{kyle2023argument} introduced a supervised ASC tagger trained on a treebank that integrates semantic information across key construction types. Their system targeted nine ASC types, each defined by a characteristic mapping between semantic and syntactic frames (Appendix \ref{ape:A}).

Building on this supervised training approach and the selected ASC types, \citet{sung2024leveraging} evaluated multiple training strategies and found that a RoBERTa-based tagger trained on a combined L1 and L2 gold-standard treebank \citep{sung2024annotation} achieved high F1 scores across L2 writing (0.915) and L2 speaking (0.928) domains. The results suggest that the tagger is reasonably robust across L2 production modes, providing a foundation for downstream tools that compute ASC-based indices for corpus-based L2 proficiency analysis.

\begin{figure}[!]
  \centering
  \includegraphics[width=0.49\textwidth]{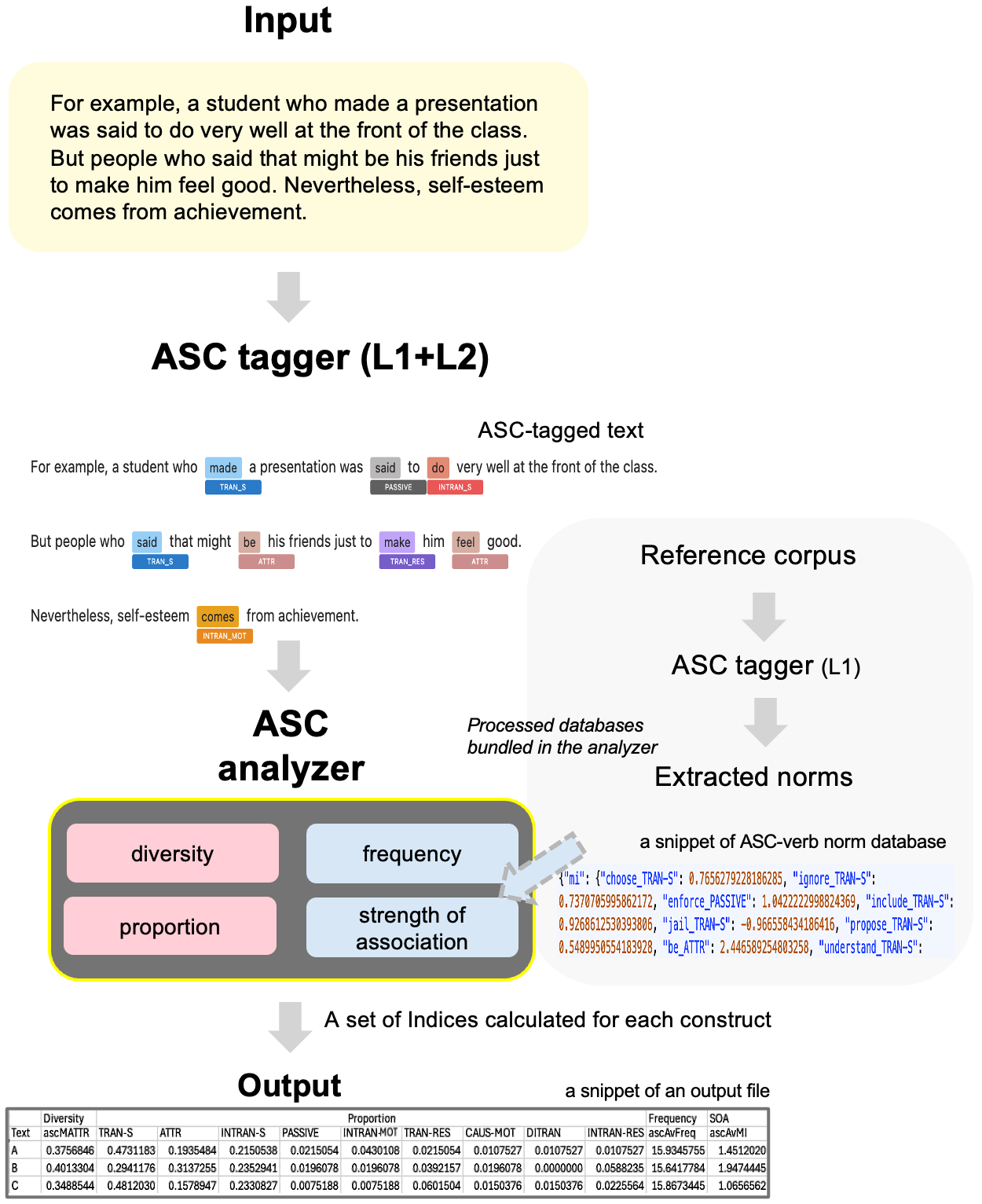} 
  \caption{High-level architecture of ASC analyzer}
  \label{fig:1}
\end{figure}

\section{ASC analyzer architecture}
ASC analyzer is designed to compute interpretable indices that quantify the use of ASCs in English texts. Based on ASC annotations generated by the ASC tagger, the analyzer transforms these labels into a set of operational metrics.

As illustrated in Figure~\ref{fig:1}, the analyzer processes ASC-tagged output from input texts and calculates four families of indices. Diversity and proportion are text-internal measures that reflect the range and distribution of ASC types and ASC-verb lemma pair types within each text. In contrast, frequency and strength of association (SOA) are text-external measures, computed by comparing ASC usage in the input texts to norms from reference corpora, capturing how often input texts include common or strongly associated ASC-verb combinations. Note that based on the F1 scores reported in \citet{sung2024leveraging}, the Gold L1+L2 model was used to process input texts due to its higher accuracy in L2 contexts, while the Gold L1 model was applied to the reference corpus for its more stable performance in L1 contexts. See Appendix \ref{ape:B} for detailed F1 scores of each tagger.

\subsection{ASC‑based Indices}
Below we formalize each index family. Implementations in Python follow the equations verbatim.

\paragraph{Diversity.} The moving‑average type-token ratio (MATTR; \citealp{covington2010cutting}) with a sliding window \(w\) (default: 11) for a token sequence \(X\) of length \(N\) is defined as:
\begin{equation*}
\resizebox{\linewidth}{!}{$
\begin{split}
\mathrm{MATTR}_{w}(X) &= 
  \frac{1}{N-w+1}
  \sum_{i=1}^{N-w+1}
    \frac{\lvert \mathrm{types}(X_{i:i+w-1})\rvert}{w},\\
&\quad\text{if }N\ge w+1,
\end{split}
$}
\end{equation*}
We derive three variants: \ \(\mathit{ascMATTR}\) (ASC tokens), \ \(\mathit{ascLemmaMATTR}\) (ASC-verb lemma pairs), and \ \(\mathit{ascLemmaMATTRNoBe}\) (ASC-verb lemma pairs excluding \textit{be}).

\paragraph{Proportion.} 
For each construction type \(c\), we define its proportion (\citealp{hwang2023automatic}) in the text as
\begin{equation*}
  \mathrm{Prop}_{c}(X)
  = \frac{f_{c}}{N_{\mathrm{ASC}}},
  \label{eq:prop}
\end{equation*}
where \(f_{c}\) is the number of tokens of type \(c\) in \(X\), and \(N_{\mathrm{ASC}}\) is the total number of ASC tokens in \(X\). This yields nine variants, one per ASC type (e.g.,  \ \(\mathit{ATTR\_Prop}\)).

\paragraph{Frequency.}  
Let an input text contain \(M\) tokens \(t_{1},\dots,t_{M}\), where each \(t_{i}\) is matched up to its raw frequency \(f^{\mathrm{ref}}(t_{i})\) in a reference corpus (excluding types with \(f^{\mathrm{ref}}<5\)).  Defining  
\[
\ell(t_{i}) = \ln\bigl(f^{\mathrm{ref}}(t_{i})\bigr),
\]  
we compute a frequency index as:  
\begin{equation*}
  \mathrm{Freq}
  = \frac{1}{M}
    \sum_{i=1}^{M} \ell(t_{i}).
  \label{eq:freq}
\end{equation*}
Two variants are derived, depending on the selected token sets: \ \(\mathit{ascAvFreq}\) (ASC tokens) and \ \(\mathit{ascLemmaAvFreq}\) (ASC verb‐lemma pairs), following the approach of \citet{kyle2017assessing}.

\paragraph{SOA.}
For each ASC–verb lemma pair \((c, v)\), SOA scores are computed from frequency counts in a reference corpus, where \(a = f_{c,v}\), \(b = f_{\bar{c},v}\), \(c = f_{c,\bar{v}}\), and \(d = f_{\bar{c},\bar{v}}\), with total corpus size \(N = a + b + c + d\). The expected frequency of the pair is given by:
\begin{equation*}
\mathrm{E}(c, v) = \frac{(a + b)(a + c)}{N}
\end{equation*}

Based on these values, we define four pointwise association metrics: mutual information (MI), t-score (T), and two \(\Delta P\) values:

\begin{equation*}
\mathrm{MI}(c, v) = \log_2 \left( \frac{a}{\mathrm{E}(c, v)} \right)
\end{equation*}

\begin{equation*}
\mathrm{T}(c, v) = \frac{a - \mathrm{E}(c, v)}{\sqrt{a}}
\end{equation*}

\begin{align*}
\Delta \mathrm{P}_{\text{Lemma}}(c, v) &= \frac{a}{a + b} - \frac{c}{c + d} \\
\Delta \mathrm{P}_{\text{Structure}}(c, v) &= \frac{a}{a + c} - \frac{b}{b + d}
\end{align*}

We derive two text-level SOA indices: \(\mathit{ascAv*}\), the mean score across all ASC-lemma tokens (e.g., \(\mathit{ascAvMI}\)), and \(\mathit{t*}\), a type-specific mean computed only over tokens labeled with ASC type \(t\) (e.g., \(\mathit{ATTR\_AvMI}\)). This indexing approach follows \citet{gries2015statistical} and \citet{kyle2017assessing}.

\subsection{Reference corpora for norm extraction}
\label{sec:refcorpora}
As briefly explained, frequency and SOA are text-external measures that reflect how closely an input text aligns with constructional norms from large external corpora. In its current version, the analyzer draws on two reference corpora:

\paragraph{\texttt{cow}}
We used a subset of the English Corpus of the Web
(EnCOW; \citealp{schafer2015processing, schafer2012building}). It contains 
\mbox{360{,}783{,}433} tokens,
15{,}439{,}673 sentences, and
39{,}838{,}785 automatically tagged ASCs.

\paragraph{\texttt{subt}}
We used the SUBTLEX-US corpus
of American film and television subtitles
(\citealp{brysbaert2012adding,brysbaert2009moving}).
The version used here comprises
\mbox{76{,}965{,}430} tokens,
164{,}686 word types,
5{,}128{,}462 sentences, and
5{,}665{,}251 tagged ASCs across 8{,}388 subtitle files.

\section{Using ASC analyzer: From installation to application}
\subsection{Installation and quick start}

First, install the required dependencies and the ASC analyzer package:
\begin{lstlisting}[language=bash]
pip install spacy
pip install spacy-transformers
python -m spacy download en_core_web_trf
pip install asc-analyzer
\end{lstlisting}

Next, view the available options:
\begin{lstlisting}
python3 -m asc_analyzer.cli --help
\end{lstlisting}

To analyze a directory of input texts and save the features to CSV, run:
\begin{lstlisting}
python3 -m asc_analyzer.cli \
  --input-dir "/path/to/texts" \
  --output-csv "/path/to/output.csv" \
  --source "cow"    # or "subt"
\end{lstlisting}

\subsection{Application: ELLIPSE Corpus}
\label{sec:ellipse}

To demonstrate the utility of the ASC analyzer in L2 research, we conducted both bivariate and multivariate analyses using a large-scale ESL writing dataset. We used 6,482 essays from the ELLIPSE corpus \citep{crossley2023english}, a reliability-filtered subset of U.S. statewide writing assessments spanning grades 8–12 across 29 prompts. Each essay includes six analytic scores for cohesion, syntax, vocabulary, phraseology, grammar, and conventions. To construct a composite proficiency index, we averaged the four subscores most aligned with constructional usage (syntax, vocabulary, phraseology, and grammar). The constructional norms were derived from the \texttt{COW} to compute frequency and SOA indices.

\subsection{Modeling the relationship between ASC use and L2 writing proficiency}

\paragraph{Bivariate correlations:}  
Pearson correlations were computed between each ASC-based index and the composite writing score, retaining only those with \(|r|\ge0.10\) \citep{cohen2013statistical}.\footnote{\label{fn:soa}Within each SOA family, we retained only the index most strongly correlated with the scores.} As shown in Table~\ref{tab:1}, \(\mathit{ascMATTR}\) yielded the strongest positive correlation (\(r = 0.26\)), while the frequency-based index \(\mathit{ascAvFreq}\) showed the strongest negative correlation (\(r = -0.22\)). Although the correlations were modest overall, the results align with previous findings: more proficient L2 writers tend to use a wider variety of ASC types \citep{hwang2023automatic} and rely less on highly frequent, but strongly entrenched, verb–construction pairings—except in the case of simple transitives \citep{kyle2017assessing}.

\begin{table}[H]
  \resizebox{0.8\columnwidth}{!}{%
  \begin{tabular}{llr}
    \toprule
    Construct & Index & $r$ \\
    \midrule
    Diversity &
      ascMATTR              & \textbf{.26} \\
    & ascLemmaMATTR        & .16 \\
    & ascLemmaMATTRNoBe    &  .11 \\
    \addlinespace
    Proportion &
      ATTR\_Prop                   & -.11 \\
    & CAUS.MOT\_Prop              &  .06 \\
    & DITRAN\_Prop                &  .06 \\
    & INTRAN.MOT\_Prop            &  .04 \\
    & INTRAN.RES\_Prop            & .13 \\
    & INTRAN.S\_Prop              &  .06 \\
    & PASSIVE\_Prop               & .19 \\
    & TRAN.RES\_Prop              & .16 \\
    & TRAN.S\_Prop                & -.13 \\
    \addlinespace
    Frequency &
      ascAvFreq                    & \textbf{-.22} \\
    & ascLemmaAvFreq              & -.15 \\
    \addlinespace
    SOA &
      asc\_AvMI                    &  .12 \\
    & CAUS.MOT\_AvMI              &  .09 \\
    & DITRAN\_AvMI                &  .11 \\
    & INTRAN.MOT\_AvMI            &  .08 \\
    & INTRAN.RES\_AvMI            & .20 \\
    & INTRAN.S\_AvMI              &  .11 \\
    & PASSIVE\_AvMI               & .15 \\
    & TRAN.RES\_AvMI              & .14 \\
    & TRAN.S\_\(\Delta \mathrm{P}_{\text{Structure}}\) & -.14 \\
    \bottomrule
  \end{tabular}}
  \caption{Correlations between ASC-based indices and L2 writing scores}
  \label{tab:1}
\end{table}

\paragraph{Multivariate regression:}
Indices that passed the bivariate filter were entered into an AIC-based model selection procedure (\(\Delta\mathrm{AIC}<4\); \citealp{akaike2003new, tan2012reliability}), with multicollinearity controlled beforehand. The final model retained 12 ASC-based predictors and explained a modest proportion of variance in writing scores (\(R^2_{\text{adj}} = 0.143\); Table~\ref{tab:2}), with an overall correlation of \(r \approx 0.38\).

\begin{table}[H]
  \centering
  \resizebox{0.45\textwidth}{!}{%
  \begin{tabular}{lrrrrr}
    \toprule
    Predictor & Estimate & SE & $t$ & $p$ & Rel.\ Imp.\ (\%) \\
    \midrule
    Intercept                    &  3.001 & 0.106 & 28.26 & $<$.001 & --   \\
    ascMATTR            &  0.892 & 0.204 &  4.37 & $<$.001 & 16.4 \\
    ATTR\_Prop          & -0.902 & 0.106 & -8.48 & $<$.001 &  8.1 \\
    DITRAN\_AvMI        &  0.013 & 0.003 &  4.58 & $<$.001 &  4.2 \\
    INTRAN.RES\_AvMI    &  0.036 & 0.004 &  9.95 & $<$.001 & 15.2 \\
    INTRAN.RES\_Prop    & -1.082 & 0.502 & -2.15 &  .031   &  3.4 \\
    INTRAN.S\_AvMI      &  0.052 & 0.007 &  7.32 & $<$.001 &  6.1 \\
    PASSIVE\_AvMI       &  0.052 & 0.006 &  8.10 & $<$.001 &  9.2 \\
    PASSIVE\_Prop       &  2.242 & 0.294 &  7.62 & $<$.001 & 12.2 \\
    TRAN.RES\_AvMI      &  0.023 & 0.005 &  5.09 & $<$.001 &  5.7 \\
    TRAN.RES\_Prop      &  0.650 & 0.240 &  2.71 &  .007   &  5.9 \\
    TRAN.S\_\(\Delta \mathrm{P}_{\text{Structure}}\) & -8.372 & 1.174 & -7.13 & $<$.001 &  7.8 \\
    TRAN.S\_Prop        & -0.518 & 0.101 & -5.14 & $<$.001 &  5.7 \\
    \midrule
    \multicolumn{6}{l}{
      $R^{2}=0.145$ \,(adj.\ $0.143$); RSE $=0.521$; 
      $F(12,6469)=91.1$, $p<.001$
    }\\
    \bottomrule
  \end{tabular}}
  \caption{Summary of the regression model predicting L2 writing scores}
  \label{tab:2}
\end{table}

\subsection{Comparative evaluation with other models}
\paragraph{Comparison with a syntactic complexity-based model:}
To evaluate the explained variance of ASC-based indices, we compared their predictive power against a multivariate model composed of syntactic complexity measures widely used in L1/L2 acquisition research \cite{hunt1965grammatical, lu2011corpus, ortega2003syntactic}.

Drawing from prior studies \cite{biber2016predicting, kyle2016measuring, kyle2017assessing}, syntactic complexity was operationalized using a set of text-internal indices that capture structural elaboration and grammatical maturity in learner writing. These indices were grouped into three broad categories based on the syntactic units they quantify:
(1) Unit length, which includes measures such as the mean length of clause;
(2) Clausal complexity, which captures the frequency and depth of embedded clause constructions; and
(3) Phrasal complexity, which reflects the internal modification and elaboration of noun phrases.
The full list of representative indices is provided in Appendix~\ref{ape:C}.

Following the same procedure used for the ASC-based indices, we built a regression model using the suite of syntactic complexity indices. The final model yielded a lower adjusted \(R^2 = 0.077\). This comparison suggests that, for this corpus, ASC-based indices accounted for a greater portion of the variance in L2 writing scores than models based solely on syntactic complexity measures.

\paragraph{Comparison with an alternative lexicogrammatical complexity model:} In studies of this kind, it is also important to examine whether newly proposed indices offer unique explanatory power beyond existing measures of lexicogrammatical complexity. To address this, we compared the ASC-based indices against a second baseline model composed of well-established lexicogrammatical indices, which primarily capture complexity at the word and bigram levels \cite{bulte2025complexity}.

Following prior research \cite{kyle2023assessing, paquot2018phraseological}, the lexicogrammatical indices in this study fall into three main categories: (1) Syntactic dependency bigrams, which measure the SOA between syntactically linked words (e.g., verb and object pairs); (2) Contiguous lemmatized bigrams, which capture lexical co-occurrence patterns independent of syntactic structure; and (3) Word-level indices, reflecting lexical sophistication (e.g., frequency, concreteness, contextual and associative distinctiveness) and diversity. Full descriptions of these indices are provided in Appendix~\ref{ape:D}.

The baseline model, which included only word- and phrase-level lexicogrammatical indices, achieved an adjusted \( R^2 = 0.363 \), while the combined model incorporating both lexicogrammatical and ASC-based indices yielded a higher adjusted \( R^2 = 0.390 \), reflecting an increase of \leavevmode\( \Delta R^2 = 0.027 \). This result suggests that ASC-based indices may capture an additional variance in L2 writing scores, offering complementary insights into constructional aspects of language use not fully accounted for by existing lexicogrammatical measures.

\section{Conclusion}

This study introduced the ASC analyzer, an open-source toolkit designed for L2 researchers and applied linguists interested in examining ASC usage in English texts. Through a proof-of-concept analysis, we demonstrated how ASC-based indices can quantify constructional patterns in L2 writing and examined their relationships with writing proficiency scores. We also compared their explanatory power against traditional syntactic complexity indices and assessed how much additional variance they capture beyond existing lexicogrammatical measures. Additional information about the package is available at \url{https://github.com/hksung/ASC-analyzer}.

\section*{Limitations}

Several limitations should be acknowledged. First, the outputs of the ASC tagger may be influenced by model-internal biases and training data limitations, which can affect the accuracy and reliability of the extracted indices. As one reviewer noted, certain ASC types (e.g., intransitive resultatives) were underrepresented in the training data, potentially limiting performance for these low-frequency but pedagogically relevant constructions.

Second, the constructional norms used to calculate frequency and SOA scores were derived from a limited set of reference corpora. While these corpora provide useful native-speaker baselines, they may not fully capture the range of registers or genres present in the target texts.

Third, as proof of concept, this work focused on modeling rather than interpretation and did not conduct a detailed linguistic analysis of ASC usage.

\section*{Acknowledgments}
This research was supported by the Harold Gulliksen Psychometric Research Fellowship (2024–2025) at ETS.

\bibliographystyle{acl_natbib}
\bibliography{anthology,acl2021}

\appendix

\clearpage
\newpage 
\onecolumn
\section{Target ASCs and semantic-syntactic representations}
\label{ape:A}
This table is reproduced from Table 1 in \citet{sung2024annotation}.
\begin{table}[hbp]
\resizebox{0.9\textwidth}{!}{%
\begin{tabular}{lll}
\toprule
\textbf{ASC (Tag)} & \textbf{Semantic frame} & \textbf{Syntactic frame} \\
\midrule
Attributive (ATTR) & theme–VERB–attribute & nsubj–cop–root \\
Caused-motion (CAUS\_MOT) & agent–VERB–theme–destination & nsubj–root–obj–obl \\
Ditransitive (DITRAN) & agent–VERB–recipient–theme & nsubj–root–iobj–obj \\
Intransitive motion (INTRAN\_MOT) & theme–VERB–goal & nsubj–root–obl \\
Intransitive simple (INTRAN\_S) & agent–VERB & nsubj–root \\
Intransitive resultative (INTRAN\_RES) & theme–VERB–result & nsubj–root–advmod \\
Passive (PASSIVE) & theme–aux–V\textsubscript{passive} & nsubj:pass–aux:pass–root \\
Transitive simple (TRAN\_S) & agent–VERB–theme & nsubj–root–obj \\
Transitive resultative (TRAN\_RES) & agent–VERB–theme–result & nsubj–root–obj–xcomp \\
\bottomrule
\end{tabular}}
\end{table}

\section{F1 scores across ASC types by model and domain}
\label{ape:B}

This table, adapted from Table 2 in \citet{sung2024leveraging}, reports F1 scores by ASC tag across two taggers: one trained only on the L1 treebank (Gold L1) and another trained on a combined L1+L2 treebank (Gold L1+L2). Each model is evaluated on three test sets (L1, L2 writing, and L2 speaking) to assess cross-domain robustness.

\begin{table}[h]
\resizebox{0.8\textwidth}{!}{%
\begin{tabular}{lccc|ccc}
\toprule
\multirow{2}{*}{\textbf{ASC Tag}} & \multicolumn{3}{c|}{\textbf{Gold L1}} & \multicolumn{3}{c}{\textbf{Gold L1+L2}} \\
 & \textbf{L1} & \textbf{L2-writing} & \textbf{L2-speaking} & \textbf{L1} & \textbf{L2-writing} & \textbf{L2-speaking} \\
\midrule
ATTR & 0.972 & 0.954 & 0.986 & 0.968 & 0.971 & 0.988 \\
CAUS\_MOT & 0.818 & 0.833 & 0.710 & 0.857 & 0.867 & 0.710 \\
DITRAN & 0.919 & 0.914 & 0.842 & 0.865 & 0.881 & 0.947 \\
INTRAN\_MOT & 0.800 & 0.770 & 0.789 & 0.772 & 0.807 & 0.843 \\
INTRAN\_RES & 0.750 & 0.788 & 0.800 & 0.625 & 0.813 & 0.833 \\
INTRAN\_S & 0.779 & 0.806 & 0.817 & 0.808 & 0.803 & 0.865 \\
PASSIVE & 0.920 & 0.775 & 0.938 & 0.940 & 0.865 & 0.909 \\
TRAN\_RES & 0.884 & 0.800 & 0.625 & 0.881 & 0.792 & 0.625 \\
TRAN\_S & 0.931 & 0.929 & 0.927 & 0.936 & 0.943 & 0.948 \\
\midrule
Weighted Avg. & 0.908 & 0.900 & 0.905 & 0.912 & 0.915 & 0.928 \\
\bottomrule
\end{tabular}}
\end{table}

\clearpage
\newpage

\section{Syntactic complexity indices}
\label{ape:C}
\begin{table}[h]
\resizebox{0.8\textwidth}{!}{%
\begin{tabular}{lll}
\toprule
\textbf{Dimension} & \textbf{Index} & \textbf{Description} \\
\midrule
\textbf{Clause} 
& \texttt{mlc} & Average number of words per finite clause \\
& \texttt{mltu} & Average number of words per T-unit \\
& \texttt{dc\_c} & Number of dependent clauses per clause \\
& \texttt{ccomp\_c} & Frequency of finite complement clauses \\
& \texttt{relcl\_c} & Frequency of relative clauses per clause \\
& \texttt{infinitive\_prop} & Proportion of ``to + verb'' constructions \\
& \texttt{nonfinite\_prop} & Proportion of nonfinite (gerund/participial) clauses \\
\addlinespace
\textbf{Phrase} 
& \texttt{mean\_nominal\_deps} & Average number of nominal dependents per noun \\
& \texttt{relcl\_nominal} & Relative clauses modifying nominals \\
& \texttt{amod\_nominal} & Adjectival modifiers of nominals \\
& \texttt{det\_nominal} & Determiners modifying nominals \\
& \texttt{prep\_nominal} & Prepositional phrases modifying nominals \\
& \texttt{poss\_nominal} & Possessive modifiers of nominals \\
& \texttt{cc\_nominal} & Coordinating conjunctions in noun phrases \\
\bottomrule
\end{tabular}}
\end{table}

\section{Word and bigram-level lexicogrammatical indices}
\label{ape:D}

\begin{table}[h]
\resizebox{0.9\textwidth}{!}{%
\begin{tabular}{lll}
\toprule
\textbf{Dimension} & \textbf{Index} & \textbf{Description} \\
\midrule
\textbf{Bigram} 
& \texttt{n\_amod\_\{T, MI, MI2, DP*\}} & SOA scores for noun–adjective dependencies \\
& \texttt{v\_advmod\_\{T, MI, MI2, DP*\}} & SOA scores for verb–adverb dependencies \\
& \texttt{v\_dobj\_\{T, MI, MI2, DP*\}} & SOA scores for verb–object dependencies \\
& \texttt{v\_nsubj\_\{T, MI, MI2, DP*\}} & SOA scores for verb–subject dependencies \\
& \texttt{lemma\_bg\_\{T, MI, MI2, DP*\}} & SOA scores for lemmatized word bigrams \\
\addlinespace
\textbf{Word} 
& \texttt{amod\_freq\_log} & Log frequency of adjectives \\
& \texttt{advmod\_freq\_log} & Log frequency of adverbs \\
& \texttt{adv\_manner\_freq\_log} & Log frequency of manner adverbs \\
& \texttt{mverb\_freq\_log} & Log frequency of main verbs \\
& \texttt{lex\_mverb\_freq\_log} & Log frequency of lexical main verbs \\
& \texttt{noun\_freq\_log} & Log frequency of nouns \\
& \texttt{cw\_lemma\_freq\_log} & Log frequency of content word lemmas \\
& \texttt{b\_concreteness} & Word concreteness ratings \\
& \texttt{mcd} & Contextual distinctiveness (entropy-based) \\
& \texttt{usf} & Associative distinctiveness (from USF norms) \\
& \texttt{MATTR\_11} & Moving-average type–token ratio (window = 11) \\
\bottomrule
\end{tabular}}
\caption*{\textit{Note.} DP* indicates various types of \(\Delta P\) scores, computed using either the left or right word as the cue (or the head or dependent in the case of dependency bigrams). All scores were calculated, and for the regression model, only the score showing the strongest relationship with scores was included—consistent with the treatment of the SOA scores in the baseline model (see Footnote~\ref{fn:soa}).}

\end{table}

\end{document}